\documentclass{article}
\pdfoutput=1
 \PassOptionsToPackage{numbers, compress}{natbib}
\usepackage[final]{nips_2016}
\usepackage[utf8]{inputenc} 
\usepackage[T1]{fontenc}    
\usepackage{wrapfig}

\usepackage{tikz}

\usepackage{graphicx,url,subfigure,booktabs,amsfonts,amssymb,amsmath,amsthm,nicefrac,microtype,color,bm,hyperref}

\usepackage{algorithm}
\usepackage{algpseudocode}

\newcommand{\States}{\ensuremath\mathcal{S}}
\newcommand{\Actions}{\ensuremath\mathcal{A}}
\newcommand{\Transition}{\ensuremath\mathcal{T}}
\newcommand{\Reward}{\ensuremath\mathcal{R}}
\newcommand{\Agent}{\ensuremath L}
\newcommand{\ProtocolProgram}{\ensuremath P}
\newcommand{\Human}{\ensuremath H}
\newcommand{\Environment}{\ensuremath M}
\newcommand{\HumanInput}{\ensuremath X_{\mathrm{in}}}
\newcommand{\HumanOutput}{\ensuremath X_{\mathrm{out}}}

\title{Agent-Agnostic Human-in-the-Loop \\ Reinforcement Learning}

\author{
  David Abel \\ Brown University \\ \texttt{david\_abel@brown.edu}
  \And John Salvatier \\ AI Impacts \\ \texttt{jsalvatier@gmail.com}
  \And Andreas Stuhlm\"uller \\ Stanford University \\ \texttt{andreas@stuhlmueller.org}
  \And Owain Evans \\ University of Oxford \\ \texttt{owaine@gmail.com}
}

\begin{document}

\maketitle


\begin{abstract}
Providing Reinforcement Learning agents with expert advice can dramatically improve various aspects of learning.
Prior work has developed teaching protocols that enable agents to learn efficiently in complex environments; many of these methods tailor the teacher's guidance to agents with a particular representation or underlying learning scheme, offering effective but specialized teaching procedures.
In this work, we explore {\it protocol programs}, an agent-agnostic schema for Human-in-the-Loop Reinforcement Learning.
Our goal is to incorporate the beneficial properties of a human teacher into Reinforcement Learning without making strong assumptions about the inner workings of the agent. We show how to represent existing approaches such as action pruning, reward shaping, and training in simulation as special cases of our schema and conduct preliminary experiments on simple domains.
  
\end{abstract}

\section{Introduction}

A central goal of Reinforcement Learning (RL) is to design agents that learn in a fully autonomous way. An engineer designs a reward function, input/output channels, and a learning algorithm. Then, apart from debugging, the engineer need not intervene during the actual learning process. Yet fully autonomous learning is often infeasible due to the complexity of real-world problems, the difficulty of specifying reward functions, and the presence of potentially dangerous outcomes that constrain exploration.

Consider a robot learning to perform household chores. Human engineers create a curriculum, moving the agent between simulation, practice environments, and real house environments. Over time, they may tweak reward functions, heuristics, sensors, and state or action representations. They may intervene directly in real-world training to prevent the robot damaging itself, destroying valuable goods, or harming people it interacts with.

In this example, humans do not just design the learning agent: they are also {\it in the loop} of the agent's learning process, as is typical for many learning systems. Self-driving cars learn with humans ready to intervene in dangerous situations. Facebook's algorithm for recommending trending news stories has humans filtering out inappropriate content \cite{facebook2016trending}. In both examples, the agent's environment is complex, non-stationary, and there are a wide range of damaging outcomes (like a traffic accident). As RL is applied to increasingly complex real-world problems, such interactive guidance will be critical to the success of these systems.

Prior literature has investigated how people can help RL agents learn more efficiently through different methods of interaction~\cite{ng1999policy,Knox2011,loftin2014learning,peng2016need,Thomaz2006,wiewiora2003principled,Judah2010,Griffith2013,Science2012,torrey2013teaching,knox2009interactively,Zhan2016,Walsh2011,Torrey2012,Maclin1996,Driessens2004}. Often, the human's role is to pass along knowledge about relevant quantities of the RL problem, like $Q$-values, action optimality, or the true reward for a particular state-action pair. This way, the person can bias exploration, prevent catastrophic outcomes, and accelerate learning.

Most existing work develops \textit{agent-specific} protocols for human interaction. That is, protocols for human interaction or advice that are designed for a specific RL algorithm (such as $Q$-learning). For instance,~\citet{Griffith2013} investigate the power of policy advice for a Bayesian $Q$-Learner. Other works assume that the states of the MDP take a particular representation, or that the action space is discrete or finite. Making explicit assumptions about the agent's learning process can enable more powerful teaching protocols that leverage insights about the learning algorithm or representation.

\subsection{Our contribution: agent-agnostic guidance of RL algorithms}

Our goal is to develop a framework for human-agent interaction that is (a) agent-agnostic and (b) can capture a wide range of ways a human can help an RL agent. Such a setting is informative of the structure of general teaching paradigms, the relationship and interplay of pre-existing teaching methods, and suggestive of new teaching methodologies, which we discuss in Section~\ref{sec:conclusion}. Additionally, approaching human-in-the-loop RL from a maximally general standpoint can help illustrate the relationship between the requisite power of a teacher and the teacher's effectiveness on learning. For instance, we demonstrate sufficient conditions on a teacher's knowledge about an environment that enable effective\footnote{By ``effective`` we mean: pruning bad actions while never pruning an optimal action. See Remark 3 (below).}  action pruning of an arbitrary agent. Results of this form can again be informative to the general structure of teaching RL agents.


We make two simplifying assumptions. First, we consider environments where the state is fully observed; that is, the learning agent interacts with a Markov Decision Process (MDP) \citep{puterman2014markov, kaelbling1996reinforcement,sutton1998reinforcement}. Second, we note that conducting experiments with an actual human in the loop creates a huge amount of work for a human, and can slow down training to an unacceptable degree. For this reason, we focus on programatic instantiations of humans-in-the-loop; a person informed about the task (MDP) in question will write a program to facilitate various teaching protocols. 

There are obvious disadvantages to agent-agnostic protocols. The agent is not specialized to the protocol, so it is unable to ask the human informative questions as in \cite{amir2016interactive}, or will not have an observation model that faithfully represents the process the human uses to generate advice, as in \cite{Griffith2013,Judah2010}. Likewise, the human cannot provide optimally informative advice to the agent as they don't know the agent's prior knowledge, exploration technique, representation, or learning method.

Conversely, agent-specific protocols may perform well for one type of algorithm or environment, but poorly on others. In many cases, without further hand-engineering, agent-specific protocols can't be adapted to a variety of agent-types. When researchers tackle challenging RL problems, they tend to explore a large space of algorithms with important structural differences: some are model-based vs. model-free, some approximate the optimal policy, others a value function, and so on. It takes substantial effort to adapt an advice protocol to each such algorithm. Moreover, as advice protocols and learning algorithms become more complex, greater modularity will help limit design complexity.

\begin{figure*}
\centering
\includegraphics[width=0.9\columnwidth]{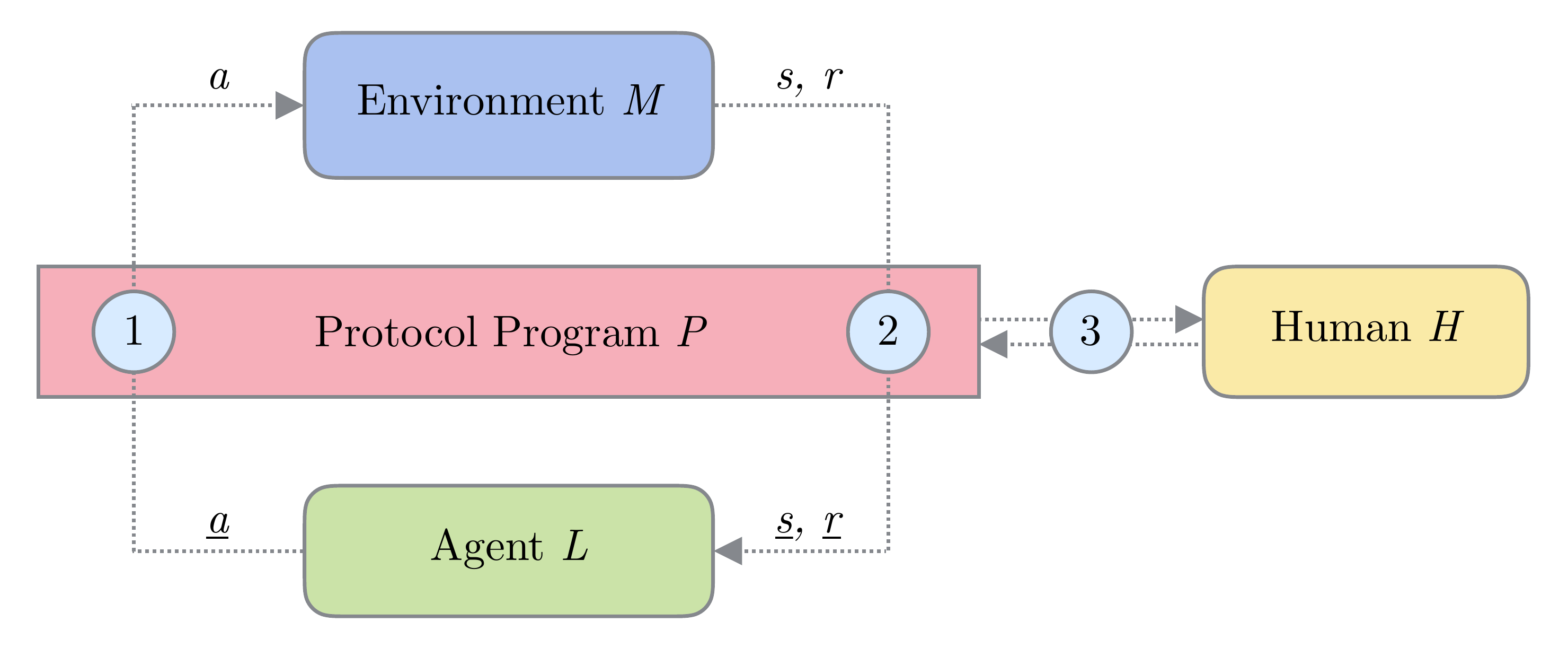}
\caption{A general setup for RL with a human in the loop. By instantiating $\ProtocolProgram$ with different protocol programs, we can implement different mechanisms for human guidance of RL agents. \label{fig:hrl}}
\vspace{-4mm}
\end{figure*}

In our framework, the interaction between the person guiding the learning process, the agent, and the environment is formalized as a {\em protocol program}. This program controls the \textit{channels} between the agent and the environment based on human input, pictured in Figure \ref{fig:hrl}. This gives the teacher extensive control over the agent: in an extreme case, the agent can be prevented from interacting with the real environment entirely and only interact with a simulation. At the same time, we require that the human \textit{only} interact with the agent during learning through the protocol program---both agent and environment are a black box to the human.


\newpage
\section{Framework}
\label{sec:frame}

Any system for RL with a human in the loop has to coordinate three components:

\begin{enumerate}
\item The {\em environment} is an MDP and is specified by a tuple $\Environment = (\States,\Actions,\Transition,\Reward,\gamma)$, where $\States$ is the state space, $\Actions$ is the action space, $\Transition \colon \States \times \Actions \times \States \mapsto [0,1]$, denotes the transition function, a probability distribution on states given a state and action, $\Reward \colon \States \times \Actions \mapsto \mathbb{R}$ is the reward function, and $\gamma$ is the discount factor.
\item The {\em agent} is a (stateful, potentially stochastic) function $\Agent \colon \States \times \Reward \rightarrow \Actions$.
\item The {\em human} can receive and send advice information of flexible type, say $\HumanInput$ and $\HumanOutput$, so, we will treat the human as a (stateful, potentially stochastic) function $\Human \colon \HumanInput \rightarrow \HumanOutput$. For example, $\HumanInput$ might contain the history of actions, states, and rewards so far, and a new proposed action $a'$, and $\HumanOutput$ might be an action as well, either equivalent to $a'$ (if accepted) or different (if rejected). We assume that the human knows in general terms how their responses will be used and is making a good-faith effort to be helpful.
\end{enumerate}

The interaction between the environment, the agent, and a human advisor sets up a mechanism design problem: how can we design an interface that orchestrates the interaction between these components such that the combined system maximizes the expected sum of $\gamma$-discounted rewards from the environment? In other words, how can we write a protocol program $\ProtocolProgram \colon \States \times \Reward \rightarrow \Actions$ that can take the place of a given agent $\Agent$, but that achieves higher rewards by making efficient use of information gained through sub-calls to $\Agent$ and $\Human$?

By formalizing existing and new techniques as programs, we facilitate understanding and comparison of these techniques within a common framework. By abstracting from particular agents and environments, we may better understand the mechanisms underlying effective teaching for Reinforcement Learning by developing portable and modular teaching methods.

\section{Capturing Existing Advice Schemes}

Naturally, protocol programs cannot capture {\em all} advice protocols. Any protocol that depends on prior knowledge of the agent's learning algorithm, representation, priors, or hyperparameters is ruled out. Despite this constraint, the framework can capture a range of existing protocols where a human-in-the-loop guides an agent. 

Figure \ref{fig:hrl} shows that the human can manipulate the \textit{actions} ($\Actions$) sent to the environment, the agent's observed \textit{states} ($\States$), and observed \textit{rewards} ($\Reward$). This points to the following combinatorial set of protocol families in which the human manipulates one or more of these components to influence learning: \[\{\States,\Actions,\Reward, (\States, \Actions), (\States, \Reward), (\Actions, \Reward), (\States, \Actions, \Reward)\}\]

The first three elements of the set correspond to \textit{state manipulation}, \textit{action pruning}, and \textit{reward shaping} protocol families.\footnote{State manipulation can correspond to abstraction or training in simulation} The remaining elements represent families of teaching schemes that modify multiple elements of the agent's learning; these protocols may introduce powerful interplay between the different components, which hope future work will explore.

We now demonstrate simple ways in which protocol programs instantiate typical methods for intervening in an agent's learning process.

\begin{figure}[h]
 \begin{minipage}[t]{.48\textwidth}
 \setlength{\intextsep}{.5\baselineskip}  
\begin{algorithm}[H]
\caption{Agent in control (standard)} 
\begin{algorithmic}[1]
\Procedure{agentControl}{$s, r$}
  \State \Return $L(s, r)$
\EndProcedure    
\end{algorithmic}
\end{algorithm}
\begin{algorithm}[H]
\caption{Human in control} 
\begin{algorithmic}[1]
\Procedure{humanControl}{$s, r$}
  \State \Return $H(s, r)$
\EndProcedure  
\end{algorithmic}
\end{algorithm}
\begin{algorithm}[H]
\caption{Action pruning} 
\begin{algorithmic}[1]
\State $\Delta \leftarrow H.\Delta$ \Comment{To Prune: $\States \times \Actions \mapsto \{0,1\}$} 
\Procedure{pruneActions}{$s, r$}
\State $\underline{a} = L(s, r)$
\While{$\Delta(s, \underline{a})$} \Comment{If Needs Pruning}
\State $\underline{r} = H[(s, \underline{a})]$
\State $\underline{a} = L(s, \underline{r})$
\EndWhile
\State \Return $\underline{a}$
\EndProcedure
\end{algorithmic}
\label{alg:prune}
\end{algorithm}
 \end{minipage}
 \hfill
\begin{minipage}[t]{.48\textwidth}
\setlength{\intextsep}{.5\baselineskip}  
\begin{algorithm}[H]
\caption{Reward manipulation}
\begin{algorithmic}[1]
\Procedure{manipulateReward}{$s, r$}
  \State $\underline{r} = H(s, r)$
  \State \Return $L(s, \underline{r})$
\EndProcedure    
\end{algorithmic}
\label{alg:rew}
\end{algorithm}
\begin{algorithm}[H]
\caption{Training in simulation} 
\begin{algorithmic}[1]
  \State $M^* = (\States,\Actions,\mathcal{T^*},\mathcal{R^*},\gamma)$ \Comment{Simulation}
  \State $\eta = []$ \Comment{History: array of $(\States \times \Reward \times \Actions)$}
  \Procedure{trainInSimulation}{$s, r$}
  \State $\underline{s} = s$
  \State $\underline{r} = r$
  \While{$H(\eta) \neq$ ``agent is ready''}
  \State $\underline{a} = L(\underline{s}, \underline{r})$
  \State append $(\underline{s}, \underline{r}, \underline{a})$ to $\eta$
  \State $\underline{r} \sim \mathcal{R^*}(\underline{s}, \underline{a})$
  \State $\underline{s} \sim \mathcal{T^*}(\underline{s}, \underline{a})$
  \EndWhile
\State \Return $L(s, r)$
\EndProcedure
\end{algorithmic}
\label{alg:sim}
\end{algorithm}
 \end{minipage}
 \caption{Many schemes for human guidance of RL algorithms can be expressed as protocol programs. These programs have the same interface as the agent $L$, but can be safer or more efficient learners by making use of human advice $H$.}
 \vspace{-6mm}
\end{figure}

\subsection{Reward shaping}

Section~\ref{sec:frame} defined the reward function $\Reward$ as part of the MDP $M$. However, while humans generally don't design the environment, we do design reward functions. Usually the reward function is hand-coded prior to learning and must accurately assign reward values to any state the agent might reach. An alternative is to have a human generate the rewards interactively: the human observes the state and action and returns a scalar to the agent. This setup has been explored in work on \textsc{TAMER} \citep{knox2009interactively}. A similar setup (with an agent-specific protocol) was applied to robotics by \citet{daniel2014active}. It is straightforward to represent rewards that are generated interactively (or online) using protocol programs.

We now turn to other protocols in which the human manipulates rewards. These protocols assume a fixed reward function $\Reward$ that is part of the MDP $M$. 

\subsubsection{Reward shaping and Q-value initialization}
In Reward Shaping protocols, the human engineer changes the rewards given by some fixed reward function in order to influence an agent's learning. \citet{ng1999policy} introduced potential-based shaping, which shapes rewards without changing an MDP's optimal policy. In particular, each reward received by the environment is augmented by a shaping function:
\begin{equation}
F(s,a,s') = \gamma\phi(s') - \phi(s),
\end{equation}
so the agent actually receives $\underline{r} = F(s,a,s') + \Reward(s,a)$. \citet{wiewiora2003principled} showed potential shaping to be equivalent (for $Q$-learners) to a subset $Q$-value initialization under some assumptions. Further, \citet{Devlin2012} propose {\it dynamic} potential shaping functions that change over time. That is, the shaping function $F$ also takes as two time parameters, $t$ and $t'$, such that:
\begin{equation}
F(s,t,s',t') = \gamma \phi(s',t') - \phi(s,t)
\end{equation}
Where $t' > t$. Their main result is that dynamic shaping functions of this form also guarantee optimal policy invariance. Similarly, \citet{wiewiora2003principled} extend potential shaping to {\it potential-based advice} functions, which identifies a similar class of shaping functions on $(s,a)$ pairs. 

In Section~\ref{sec:theory}, we show that our Framework captures reward shaping, and consequently, a limited notion of $Q$-value initialization.



\subsection{Training in Simulation}

It is common practice to train an agent in simulation and transfer it to the real world once it performs well enough. Algorithm~\ref{alg:sim} (Figure 2) shows how to represent the process of training in simulation as a protocol program. We let $M$ represent the real-world decision problem and let $M^*$ be a simulator for $M$ that is included in the protocol program. Initially the protocol program has the agent $L$ interact with $M^*$ while the human observes the interaction. When the human decides the agent is ready, the protocol program has $L$ interact with $M$ instead.


%
\subsection{ Action Pruning }
\label{section:prune}

Action pruning is a technique for dynamically removing actions from the MDP to reduce the branching factor of the search space. Such techniques have been shown to accelerate learning and planning time~\cite{sherstov2005improving,hansen1996reinforcement,rosman2012good,abel2015goal}. In Section 5, we apply action-pruning to prevent catastrophic outcomes during exploration, a problem explored by~\citet{lipton2016combating,GarciaPolo2011,Garcia2012,Hans2008,Moldovan2012}.

Protocol programs allow action pruning to be carried out {\it interactively}. Instead of having to decide which actions to prune prior to learning, the human can wait to observe the states that are actually encountered by the agent, which may be valuable in cases where the human has limited knowledge of the environment or the agent's learning ability. In Section \ref{sec:theory}, we exhibit an agent-agnostic protocol for interactively pruning actions that preserves the optimal policy while removing some bad actions.

\begin{wrapfigure}{r}{.4\linewidth}
  \centering
\includegraphics[width=0.35\columnwidth]{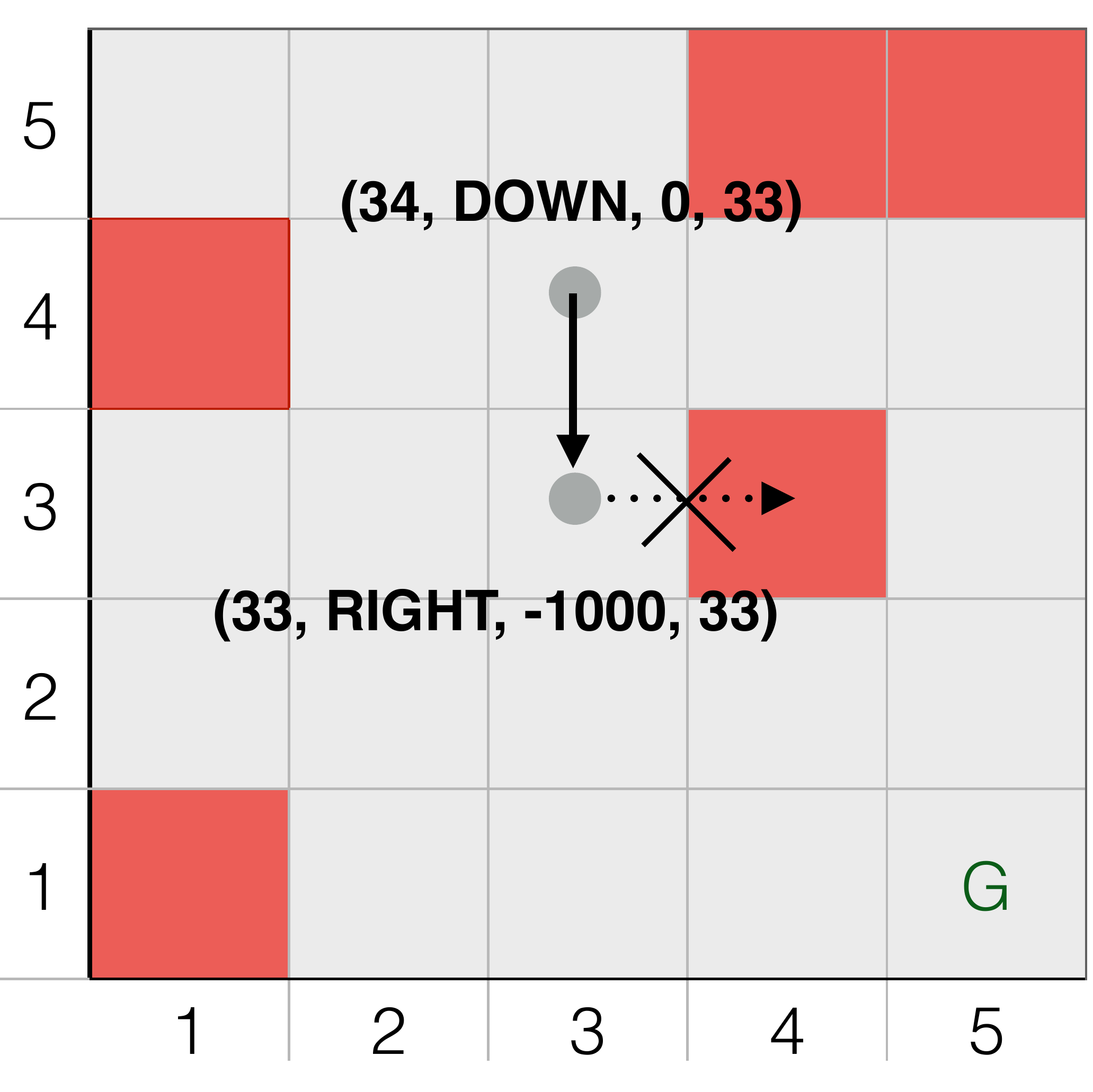}
\caption{The human allows movement from state {\bf 34} to {\bf 33} but blocks agent from falling in lava (at {\bf 43}). \label{fig:lava}}
\end{wrapfigure}

Our pruning protocol is illustrated in a gridworld with lava pits (Figure \ref{fig:lava}). The agent is represented by a gray circle, ``G'' is a goal state that provides reward +1, and the red cells are lava pits with reward $-200$. All white cells provide reward 0.

At each time step, the human checks whether the agent moves into a lava pit. If it does not (as in moving DOWN from state {\bf 34}), the agent continues as normal. If it does (as in moving RIGHT from state {\bf 33}), the human bypasses sending any action to the true MDP (preventing movement right) and sends the agent a next state of {\bf 33}. The agent doesn't actually fall in the lava but the human sends them a reward $r \leq -200$. After this negative reward, the agent is less likely to try the action again. For the protocol program, see Algorithm~\ref{alg:prune} in Figure 2.

Note that the agent receives no explicit signal that their attempted catastrophic action was blocked by the human. They observe a big negative reward and a self-loop but no information about whether the human or environment generated their observation.

\subsection{Manipulating state representation}

The agent's state representation can have a significant influence on its learning. Suppose the states of MDP $M$ consist of a number of features, defining a state vector $s$. The human engineer can specify a mapping $\phi$ such that the agent always receives $\phi(s) = \bar{s}$ in place of this vector $s$. Such mappings are used to specify high-level features of state that are important for learning, or to dynamically ignore confusing features from the agent.

This transformation of the state vector is normally fixed before learning. A protocol program can allow the human to provide processed states or high-level features interactively. By the time the human stops providing features, the agent might have learned to generate them on its own (as in Learning with Privileged Information \citep{Vapnik2009,Pechyony2010}). 

Other methods have focused on state abstraction functions to decrease learning time and preserve the quality of learned behavior, as in~\cite{li2006towards,ortner2013adaptive,even2003approximate,jong2005state,dean1997model,abelhershko2016approx,ferns2006methods}. Using a state abstraction function, agents compress representations of their environments, enabling deeper planning and lower sample complexity. Any state aggregation function can be implemented by a protocol program, perhaps dynamically induced through interaction with a teacher.


\section{Theory}
\label{sec:theory}
Here we illustrate some simple ways in which our proposed agent-agnostic interaction scheme captures other existing agent-agnostic protocols. The following results all concern Tabular MDPs, but are intended to offer intuition for high-dimensional or continuous environments as well. 

\subsection{Reward Shaping}

First we observe that protocol programs can precisely capture methods for shaping reward functions.

{\bf Remark 1}: {\it For any reward shaping function $F$, including potential-based shaping, potential-based advice, and dynamic potential-based advice, there is a protocol that produces the same rewards.}

To construct such a protocol for a given $F$, simply let the reward output by the protocol, $\underline{r}$, take on the value $F(s) + \underline{r}$ at each time step. That is, in Algorithm~\ref{alg:rew}, simply define $H(s,\underline{r}) = F(s) + \underline{r}$.

%

\subsection{Action Pruning}

We now show that there is a simple class of protocol programs that carry out action pruning of a certain form.

{\bf Remark 2}: {\it There is a protocol for pruning actions in the following sense: for any set of state action pairs $\bm{sa} \subset S \times A$, the protocol ensures that, for each pair $(s_i, a_j) \in \bm{sa}$, action $a_j$ is never executed in the MDP in state $s_i$.} \\

The protocol is as described in Section~\ref{section:prune} and shown in Algorithm~\ref{alg:prune}. The premise is this: in all cases where the agent executes an action that should be pruned, the protocol gives the agent low reward and forces the agent to self-loop.

Knowing which actions to prune is itself a challenging problem. Often, it is natural to assume that the human guiding the learner knows something about the environment of interest (such as where high rewards or catastrophes lie), but may not know every detail of the problem. Thus, we consider a case in which the human has partial (but useful) knowledge about the problem of interest, represented as an approximate $Q$-function. The next remark shows there is a protocol based on approximate knowledge with two properties: (1) it never prunes an optimal action, (2) it limits the magnitude of the agent's worst mistake:

{\bf Remark 3}: {\it Assuming the protocol designer has a $\beta$-optimal $Q$ function:
\begin{equation}
||Q^*(s,a) - Q_H(s,a) ||_\infty \leq \beta
\end{equation}
there exists a protocol that never prunes an optimal action, but prunes all actions so that the agent's mistakes are never more than $4\beta$ below optimal. That is, for all times $t$:
\begin{equation}
V^{L_t}(s_t) \geq V^*(s_t) - 4\beta,
\end{equation}
where $L_t$ is the agent's policy after $t$ timesteps.}

\renewcommand*{\proofname}{Proof of Remark 3}
\begin{proof}
The protocol designer has a $\beta$-approximate $Q$ function, denoted $Q_H$, defined as above. Consider the state-specific action pruning function $H(s)$:
\begin{equation}
H(s) = \left\{ a \in \mathcal{A} \mid Q_H(s,a) \geq \max_{a'} Q_H(s,a') - 2\beta \right\}
\end{equation}

The protocol prunes all actions not in $H(s)$ according to the self-loop method described above. This protocol induces a pruned Bellman Equation over available actions, $H(s)$, in each state:
\begin{equation}
V_H(s) = \max_{a \in H(s)} \left( \mathcal{R}(s,a) + \gamma \sum_{s'} \mathcal{T}(s,a,s') V_H(s') \right)
\end{equation}
Let $a^*$ denote the true optimal action: $a^*=\text{arg}\max_{a'} Q^*(s,a')$. To preserve the optimal policy, we need $a^* \in H(s)$, for each state. Note that $a^* \not \in H(s)$ when:
\begin{equation}
Q_H(s,a^*) < \max_{a'} Q_H(s,a') - 2\beta
\end{equation}
But by definition of $Q_H(s,a)$:
\begin{equation}
|Q_H(s,a^*) - \max_a Q_H(s,a)| \leq 2\beta
\end{equation}
Thus, $a^* \in H(s)$ can never occur. Furthermore, observe that $H(s)$ retains all actions $a$ for which:
\begin{equation}
Q_H(s,a) \geq \max_{a'} Q_H(s,a') - 2\beta,
\label{eq:qh_b}
\end{equation}
holds. Thus, in the worst case, the following two hold:
\begin{enumerate}
\item The optimal action estimate is $\beta$ too low: $Q_H(s,a^*) = Q^*(s,a^*) - \beta$
\item The action with the lowest value, $a_{bad}$, is $\beta$ too high: $Q_H(s,a_{bad}) = Q^*(s,a_{bad}) + \beta$
\end{enumerate}
From Equation~\ref{eq:qh_b}, observe that the minimal $Q^*(s,a_{bad})$ such that $a_{bad} \in H(s)$ is:
\begin{align*}
Q^*(s,a_{bad}) + \beta &\geq Q^*(s,a^*) - \beta - 2\beta \\
\therefore Q^*(s,a_{bad}) &\geq Q^*(s,a^*) - 4\beta
\end{align*}
Thus, this pruning protocol never prunes an optimal action, but prunes all actions worse then $4\beta$ below $a^*$ in value. We conclude that the agent may never execute an action $4\beta$ below optimal. \qedhere

\end{proof}

\section{Experiments}

This section applies our action pruning protocols (Section \ref{section:prune} and Remarks 2 and 3 above) to concrete RL problems. In Experiment 1, action pruning is used to prevent the agent from trying catastrophic actions, i.e. to achieve safe exploration. In Experiment 2, action pruning is used to accelerate learning. 

 \subsection{Protocol for Preventing Catastrophes}
 \label{section:catastrophe}

Human-in-the-loop RL can help prevent disastrous outcomes that result from ignorance of the environment's dynamics or of the reward function. Our goal for this experiment is to prevent the agent from taking {\em catastrophic actions}. These are real world actions so costly that we want the agent to {\em never} take the action\footnote{We allow an RL agent to take sub-optimal actions while learning. Catastrophic actions are not allowed because their cost is orders of magnitude worse than non-catastrophic actions.}. This notion of catastrophic action is closely related to ideas in ``Safe RL'' \citep{Garc2015,Moldovan2012} and to work on ``significant rare events'' \citep{paul2016alternating}.

Section \ref{section:prune} describes our protocol program for preventing catastrophes in finite MDPs using action pruning. There are two important elements of this program:
 \begin{enumerate}
 \item When the agent tries a catastrophic action $a$ in state $s$, the agent is blocked from executing the action in the real world, and the agent receives state and reward: $(s,r_{bad})$, where $r_{bad}$ is an extreme negative reward. 
 \item This $(s,a)$ is stored so that the protocol program can automate the human's intervention, which could allow the human to stop monitoring after all catastrophes have been stored.
\end{enumerate}
This protocol prevents catastrophic actions while preserving the optimal policy and having only minimal side-effects on the agent's learning. We can extend this protocol to environments with high-dimensional state spaces. Element (1) above remains the same. But (2) must be modified: preventing future catastrophes requires \textit{generalization} across catastrophic actions (as there will be infinitely many such actions). We discuss this setting in Appendix A.

\subsection{Experiment 1: Preventing Catastrophes in a Pong-like Game}
Our protocol for preventing catastrophes is intended for use in a real-world environment. Here we provide a preliminary test of our protocol in a simple video game. 

Our protocol treats the RL agent as a black box. To this end, we applied our protocol to an open-source implementation of the state-of-the-art RL algorithm ``Trust Region Policy Optimization'' from \citet{duan2016benchmarking}. The environment was {\em Catcher}, a simplified version of Pong with non-visual state representation. Since there are no catastrophic actions in Catcher, we modified the game to give a large negative reward when the paddle's speed exceeds a speed limit. We compare the performance of an agent who is assisted by the protocol (``Pruned'') and so is blocked from the catastrophic actions\footnote{
  We did not use an actual human in the loop. Instead the agent was blocked by a protocol program that checked whether each action would exceed the speed limit. This is essentially the protocol outlined in Appendix A but with the classifier trained offline to recognize catastrophes. Future work will test similar protocols using actual humans. (In this experiment a human can easily recognize catastrophic actions by reading the agent's speed directly from the game state.)} to the performance of a normal RL agent (``Not Pruned'').

\begin{figure}[!tbp]
  \centering
  \begin{minipage}[b]{0.49\textwidth}
         \includegraphics[width=\textwidth]{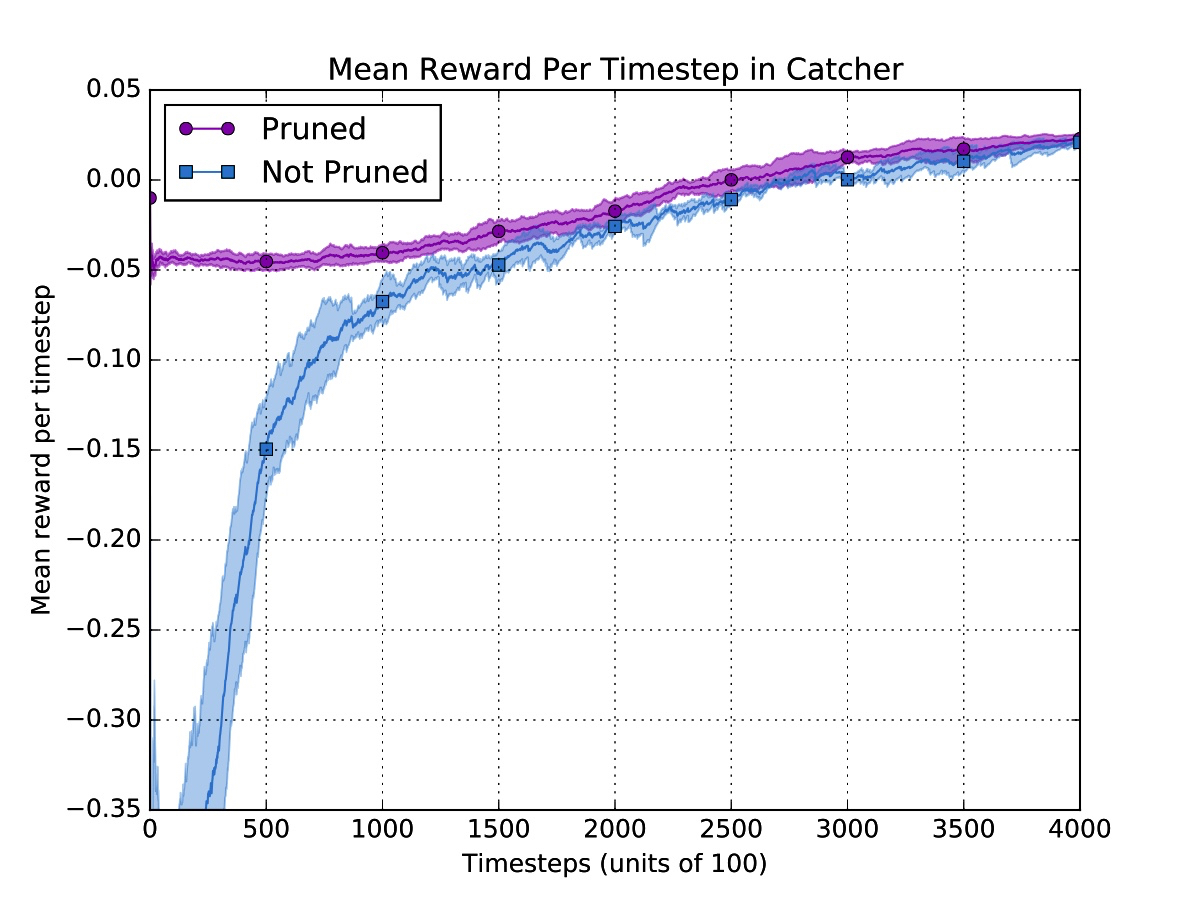}
    \caption{Preventing Catastrophic Speeds \label{fig:catch}}
  \end{minipage}
  \hfill
  \begin{minipage}[b]{0.495\textwidth}
        \vskip .1em
	\includegraphics[width=\textwidth]{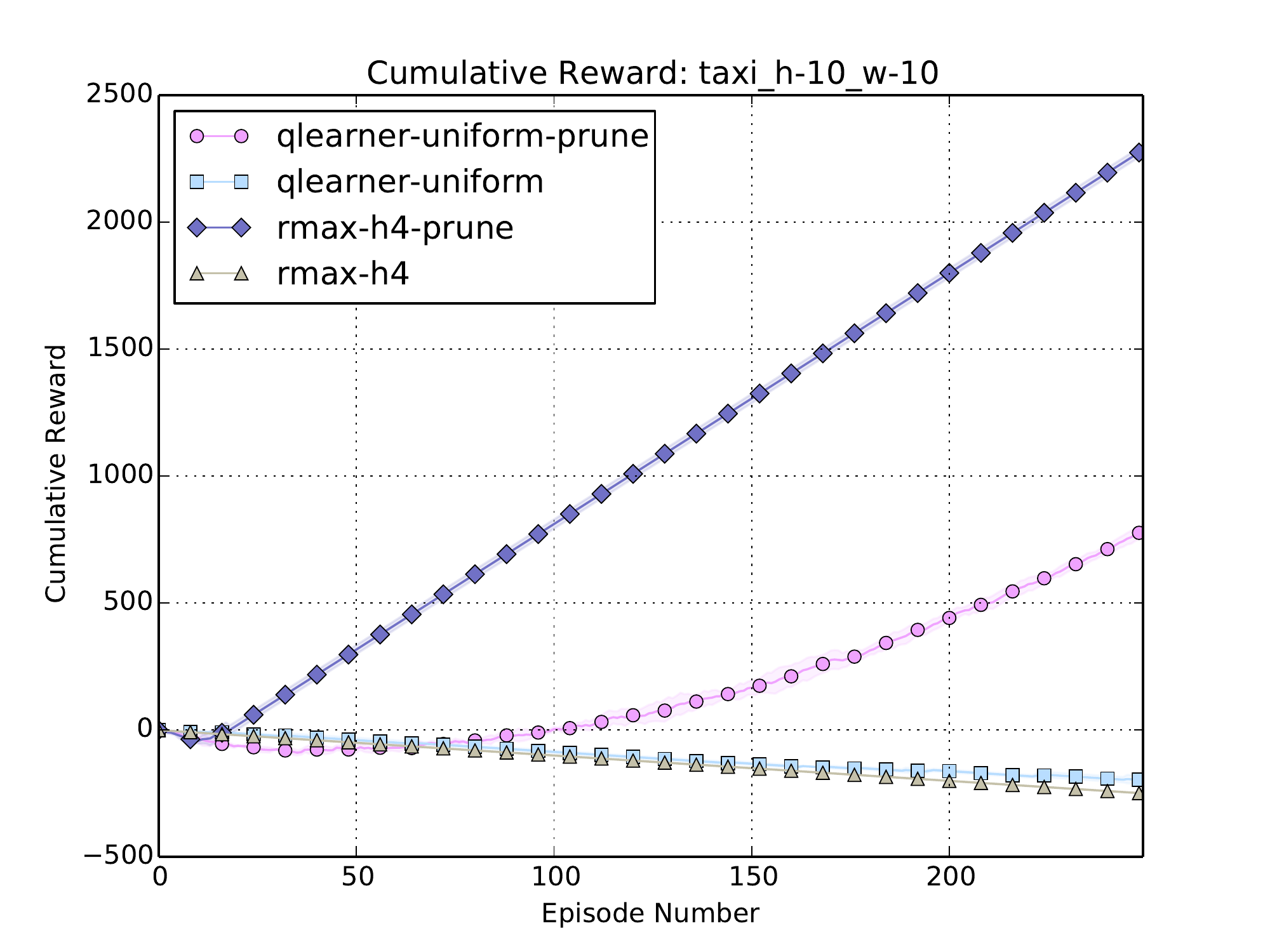}
    \caption{Pruning in Taxi. \label{fig:taxi}}
  \end{minipage}
\end{figure}

Figure~\ref{fig:catch} shows the agent's mean performance ($\pm 1$SD over 16 trials) over the course of learning. We see that the agent with protocol support (``Pruned'') performed much better overall. This is unsurprising, as it was blocked from ever doing a catastrophic action. The gap in mean performance is large early on but diminishes as the ``Not Pruned'' agent learns to avoid high speeds. By the end (i.e. after 400,000 actions), ``Not Pruned'' is close to ``Pruned'' in mean performance but its total returns over the whole period are around 5 times worse. While the ``Pruned'' agent observes incongruous state transitions due to being blocked by our protocol, Figure~\ref{fig:catch} suggests these observations do not have negative side effects on learning.

\subsection{Protocol for Accelerating Learning}

We also conducted a simple experiment in the Taxi domain from~\citet{dietterich2000hierarchical}. The Taxi problem is a more complex version of grid world: each problem instances consists of a taxi and some number of passengers. The agent directs the taxi to each passenger, picks the passenger up, and brings them to their destination and drops them off.

We use Taxi to evaluate the effect of our action pruning protocol for accelerating learning in discrete MDPs. There is a natural procedure for pruning suboptimal actions that dramatically reduces the size of the reachable state space: if the taxi is carrying a passenger but is not at the passenger's destination, we prune the \texttt{dropoff} action by returning the agent back to its current state with -0.01 reward. This prevents the agent from exploring a large portion of the state space, thus accelerating learning. 

\subsection{Experiment 2: Accelerated Learning in Taxi}
We evaluated $Q$-learning~\cite{watkins1992q} and \textsc{R-max}~\cite{brafman2003r} with and without action pruning in a simple $10\times10$ instance with one passenger. The taxi starts at $(1,1)$, the passenger at $(4,3)$ with destination $(2,2)$. We ran standard $Q$-Learning with $\varepsilon$-greedy exploration with $\varepsilon=0.2$ and with \textsc{R-max} using a planning horizon of four. Results are displayed in Figure~\ref{fig:taxi}.

Our results suggest that the action pruning protocol simplifies the problem for a $Q$-Learner and dramatically so for \textsc{R}-Max. In the allotted number of episodes, we see that pruning substantially improves the overall cumulative reward achieved; in the case of \textsc{R-max}, the agent is able to effectively solve the problem after a small number of episodes. Further, the results suggests that the agent-agnostic method of pruning is effective without having any internal access to the agent's code.

 \section{Conclusion}
\label{sec:conclusion}
We presented an agent-agnostic method for giving guidance to Reinforcement Learning agents. Protocol programs written in this framework apply to any possible RL agent, so sophisticated schemes for human-agent interaction can be designed in a modular fashion without the need for adaptation to different RL algorithms. We presented some simple theoretical results that relate our method to existing schemes for interactive RL and illustrated the power of action pruning in two toy domains.

A promising avenue for future work are dynamic state manipulation protocols, which can guide an agent's learning process by incrementally obscuring confusing features, highlighting relevant features, or simply reducing the dimensionality of the representation. Additionally, future work might investigate whether certain types of value initialization protocols can be captured by protocol programs, such as the optimistic initialization for arbitrary domains developed by ~\citet{Machado2014a}. Moreover, the full combinatoric space of learning protocols is suggestive of teaching paradigms that have yet to be explored. We hypothesize that there are powerful teaching methods that take advantage of the interplay between state manipulation, action pruning, and reward shaping. A further challenge is to extend the formalism to account for the interplay between multiple agents, in both competitive and cooperative settings. 

Additionally, in our experiments, all protocols are explicitly programmed in advance. In the future, we'd like to experiment with dynamic protocols with a human in the loop during the learning process.

Lastly, an alternate perspective on the framework is that of a {\it centaur} system: a joint Human-AI decision maker~\cite{swartout2016virtual}. Under this view, the human trains and queries the AI dynamically in cases where the human needs help. In the future, we'd like to establish and investigate formalisms relevant to the centaur view of the framework.

\newpage
\subsubsection*{Acknowledgments}
This work was supported by Future of Life Institute grant 2015-144846 and by the Future of Humanity Institute (Oxford). We thank Shimon Whiteson, James MacGlashan, and D. Ellis Herskowitz for helpful conversations. 

\bibliographystyle{plainnat}
\bibliography{arxiv}

\newpage
\appendix
\label{section:appendix}
\section{Protocol program for preventing catastrophes in high-dimensional state spaces}

We provide an informal overview of the protocol program for avoiding catastrophes. We focus on the differences between the high-dimensional case and the finite case described in Section \ref{section:catastrophe}. In the finite case, pruned actions are stored in a table. When the human is satisfied that all catastrophic actions are in the table, the human's monitoring of the agent can be fully automated by the protocol program. The human may need to be in the loop until the agent has attempted each catastrophic action once -- after that the human can ``retire''. 

In the infinite case, we replace this look-up table with a supervised classification algorithm. All visited state-actions are stored and labeled (``catastrophic'' or ``not catastrophic'') based on whether the human decides to block them. Once this labeled set is large enough to serve as a training set, the human trains the classifier and tests performance on held-out instances. If the classifier passes the test, the human can be replaced by the classifier. Otherwise the data-gathering process continues until the training set is large enough for the classifier to pass the test. 

If the class of catastrophic actions is learnable by the classifier, this protocol prevents all catastrophes and has minimal side-effects on the agent's learning. However, there are limitations of the protocol that will be the subject of future work:

\begin{itemize}

\item
The human may need to monitor the agent for a very long time to provide sufficient training data. One possible remedy is for the human to augment the training set by adding synthetically generated states to it. For example, the human might add noise to genuine states without altering their labels. Alternatively, extra training data could be sampled from an accurate generative model. 

\item
  Some catastrophic outcomes have a ``local'' cause that is easy to block. If a car moves very slowly, then it can avoid hitting an obstacle by braking at the last second. But if a car has lots of momentum, it cannot be slowed down quickly enough. In such cases a human in-the-loop would have to recognize the danger some time before the actual catastrophe would take place.

\item
  To prevent catastrophes from ever taking place, the classifier needs to correctly identify every catastrophic action. This requires strong guarantees about the generalization performance of the classifier. Yet the distribution on state-action instances is non-stationary (violating the usual i.i.d assumption). 
  
\end{itemize}

\end{document}